\pdfoutput=1
\documentclass[letterpaper, 10 pt, conference]{ieeeconf}  

\IEEEoverridecommandlockouts                              

\usepackage{graphicx}
\usepackage{epsfig} 
\usepackage{amsmath, amssymb, amsfonts} 
\usepackage{mathtools}
\usepackage[dvipsnames,table,xcdraw]{xcolor}
\usepackage{eucal}
\usepackage{indentfirst}
\usepackage{algorithm,algorithmic}
\usepackage{physics}
\usepackage{multirow}
\usepackage{multicol}
\usepackage{url}
\usepackage{cite}
\usepackage{subcaption}
\usepackage[export]{adjustbox}
\usepackage{hyperref}
\usepackage{bm}

\newcommand{\R}{\mathbb{R}}

\newcommand{\sr}{\textsc{r}}
\newcommand{\sh}{\textsc{h}}
\newcommand{\shbar}{\overline{\textsc{h}}}
\newcommand{\shmi}{\textsc{HMI}}
\newcommand{\sff}{f\!f}
\newcommand{\sfb}{f\!b}

\begin{document}
\title{\LARGE \bf
Bipedal Robot Walking Control Using Human Whole-Body Dynamic Telelocomotion 
}

\author{Guillermo Colin, Youngwoo Sim, and Joao Ramos 
\thanks{This work is supported by the National Science Foundation via grant CMMI-2043339.}
\thanks{The authors are with the Department of Mechanical Science and Engineering at the University of Illinois at Urbana-Champaign, USA.{\tt\small (gjcolin3@illinois.edu)}} 
}

\maketitle

\begin{abstract}
For humanoids to be deployed in demanding situations, such as search and rescue, highly intelligent decision making and proficient sensorimotor skill is expected. A promising solution is to leverage human prowess by interconnecting robot and human via teleoperation. Towards creating seamless operation, this paper presents a dynamic telelocomotion framework that synchronizes the gait of a human pilot with the walking of a bipedal robot. First, we introduce a method to generate a virtual human walking model from the stepping behavior of a human pilot which serves as a reference for the robot to walk. Second, the dynamics of the walking reference and robot walking are synchronized by applying forces to the human pilot and the robot to achieve dynamic similarity between the two systems. This enables the human pilot to continuously perceive and cancel any asynchrony between the walking reference and robot.  A consistent step placement strategy for the robot is derived to maintain dynamic similarity through step transitions. Using our human-machine-interface, we demonstrate that the human pilot can achieve stable and synchronous teleoperation of a simulated robot through stepping-in-place, walking, and disturbance rejection experiments. This work provides a fundamental step towards transferring human intelligence and reflexes to humanoid robots.
\end{abstract}

\section{Introduction}

Harnessing the intelligence and sensorimotor skills of humans and transferring them to humanoid robots is an idea that demands exploration. Despite remarkable progress in humanoid robot planning and control\cite{scott_atlas, grizzle_ames, MITHUMANOID, donghyun_wbc}, the reality is that autonomous humanoid robots still cannot match human capabilities to accomplish dynamic and complex whole-body locomotion and manipulation tasks. For example, during the COVID-19 pandemic transferring a disabled at-risk patient out of bed into a wheelchair requires precise whole-body manipulation while carefully regulating locomotion. Firefighting has a myriad of examples that are even more demanding such as carrying a victim to safety. 
Even worst, autonomous humanoid robot performance degrades sharply as environments become unknown, unpredictable, and hazardous. The option to await major breakthroughs in perception, sensing, artificial intelligence, and more is unreasonable. Instead, the robotics community should, in parallel, focus on leveraging human intelligence and sensorimotor control to meet today's demand for these robots. \\
\indent One approach is through human motion retargeting \cite{wholebody_retarget_Darvish, whole_body_movement_opt_retarget, whole_body_master_slave_locomotion, whole_body_master_slave_locomotion2}. These methods, which leverage human intelligence to generate whole-body motion trajectories are limited by the constraints to make the trajectories dynamically feasible. The human pilot, generating these trajectories, has no knowledge of the robot dynamics. Thus, human sensorimotor skills cannot be transferred to the robot. To address these problems researchers have studied using whole-body haptic feedback to humans for humanoid robot control \cite{balancing_feedback_vibrotactile, full_body_interface_compliant_robot_interaction}. While the results have been promising, they have been limited to semi-static tasks. There is no example of a human pilot dynamically controlling humanoid robot bipedal walking. 

The most promising work has been in \cite{Prof_TRO}. Here, the best of human motion retargeting and whole-body haptic feedback is combined. The main takeaway from this work is that if human locomotion can be abstracted, scaled to robot proportions, and modified based on knowledge of robot dynamics through whole-body haptic feedback, then dynamic synchronization between the two interconnected systems can be achieved. As shown in \cite{ProfScienceRob}, the human pilot leverages their intelligence and sensorimotor skills to balance, take steps, and even jump in synchrony with the bipedal robot, Little Hermes \cite{LittleHermes_CoDesign}. 

\begin{figure}[t]
\begin{center}
\includegraphics[width=0.9\linewidth]{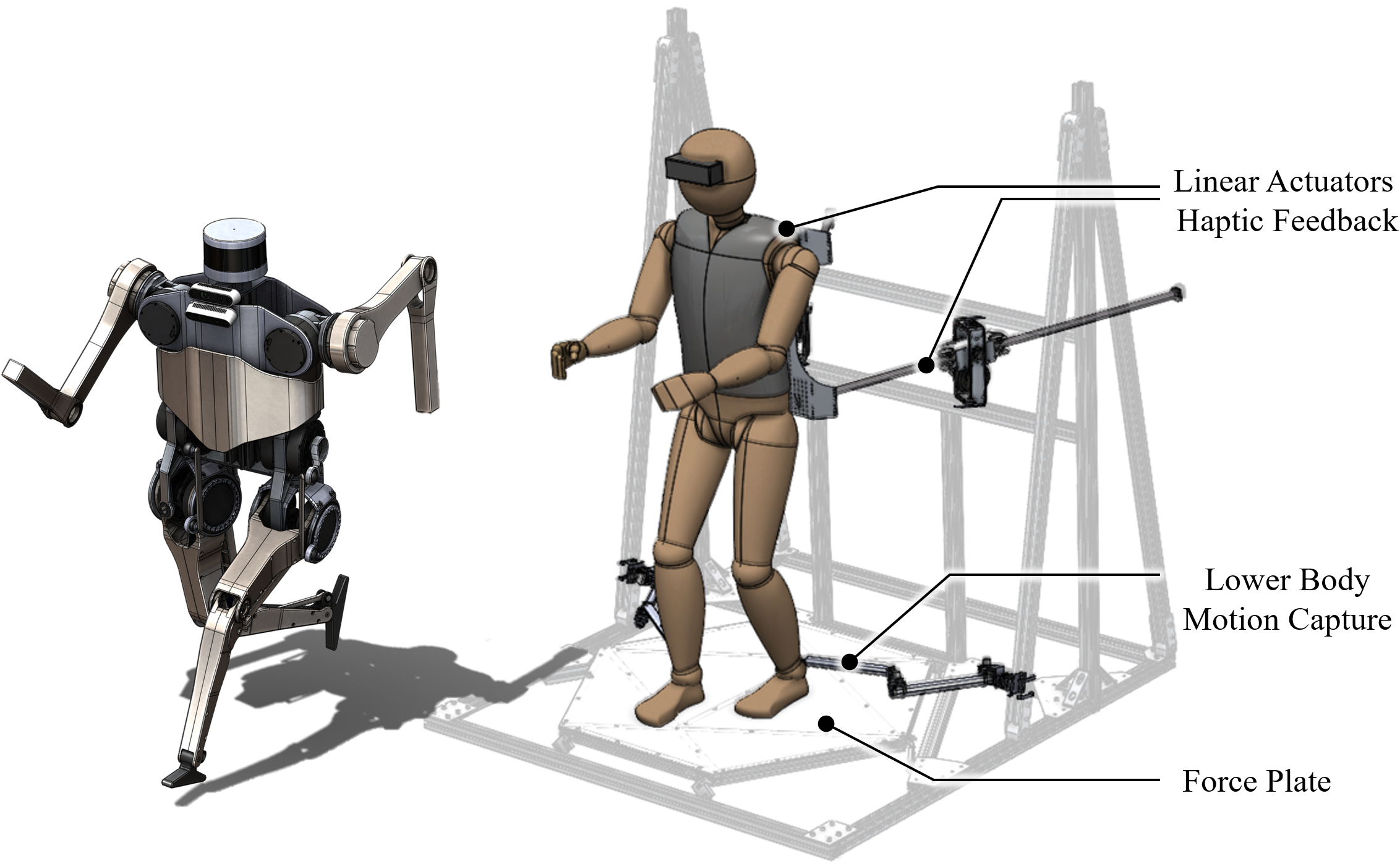}
\end{center}
\caption{Human Whole-Body Dynamic Telelocomotion Concept: The humanoid robot Tello walks via dynamic synchronization with a human pilot.}
\label{TelelocomotionConcept}
\end{figure}

In this work, we continue to explore the core idea of \cite{Prof_TRO} to achieve, to the best of our knowledge, \textit{the first synchronized bipedal walking between human and humanoid robot}. However, due to practical constraints, the abstraction of human locomotion in the frontal plane and sagittal plane has to be different. 
Our key contribution is that we introduce a method for a human pilot to intuitively generate a walking reference representing their intended walking behavior from stepping. Synchronization between the human walking reference and bipedal robot is enforced using whole-body haptic feedback and a consistent step placement law for the robot. 


\section{Preliminaries}
\label{Preliminaries}

We review three components to enable whole-body dynamic telelocomotion. First, the classical linear inverted pendulum model (LIPM) is used as a channel to scale human locomotion to the robot. Second, dynamic similarity is a concept that formalizes dynamic synchronization in this work to define equivalent motion between human pilot and robot in a normalized dynamics space \cite{ProfCartPole, Pratt_capture}. Finally, the human-machine-interface (HMI), a hardware that interconnects human and robot by enabling simultaneous motion capture and whole-body haptic force feedback to the human pilot \cite{SunyuHMI}.

\subsection{The Linear Inverted Pendulum Model}
\label{LIPMsect}

The equation of the LIPM is given by,
\begin{equation}
\label{LIPeqmain}
    \ddot{x} = \omega^{2}(x - p_{x}),
\end{equation}
where $x$ is the position of the center of mass (CoM), $\omega = \sqrt{\frac{g}{h}}$ is the natural frequency under a gravity field $g$, $h$ is the nominal height of the CoM, and $p_{x}$ is the location of the center of pressure (CoP). The LIPM contact force is derived by multiplying both sides of Eq.\,(\ref{LIPeqmain}) by the mass $m$,
\begin{equation}
\label{nonpassiveLIPforce}
    F_{x} = m\omega^{2}(x - p_{x}).
\end{equation}
This non-passive version of the LIPM allows modeling balance in bipedal systems with multiple points of contact. As seen in Eq.\,(\ref{LIPeqmain}), control of the CoP location regulates the CoM dynamics of the LIP. \\
\indent For underactuated bipedal walking, we often assume that the CoM dynamics are reduced to follow the passive LIP dynamics ($p_x = 0$) and controlled by step placement \cite{HLIP_AMBER, gong2021angular}. This model assumes that the robot has point feet and no double support phase. Hence, the only point of contact is the stance foot location. The resulting closed-form solution of the passive LIPM is the following:
\begin{equation}
\label{passiveLIPsol}
    \begin{split}
    x(t) &= c_{1}e^{\omega t} + c_{2}e^{-\omega t}, \\
    \dot{x}(t) & = \omega(c_{1}e^{\omega t} - c_{2}e^{-\omega t}),
    \end{split}
\end{equation}
where $c_{1} = \frac{1}{2}(x_{0} + \frac{1}{\omega}\dot{x}_{0})$ and $c_{2} = \frac{1}{2}(x_{0} - \frac{1}{\omega}\dot{x}_{0})$ are constants of integration consisting of the initial conditions $x_{0}$ and $\dot{x}_{0}$. \\
\indent In \cite{HLIP_AMBER} the authors define necessary and sufficient conditions for stable one-step periodic orbits (P1) for the passive LIPM. They prove that in the LIP phase portrait, both the initial and final states must lie on the orbital lines of characteristics, $\dot{x} = -\sigma_{1}x$ and $\dot{x} = \sigma_{1}x$, respectively, where the orbital slope $\sigma_1$ is defined as,
\begin{equation}
\label{orbitalslopeeq}
    \sigma_{1}:=\omega \: \coth \Big(\frac{T_{s}}{2}\omega\Big).
\end{equation}
The orbital slope is a property belonging to a family of unique P1 orbits that share the same step time duration, $T_{s}$, but different step length $\ell$. \\

\subsection{Dynamic Similarity of LIPMs}
\label{dynsimsubsect}

The motions of two dynamical systems are said to be \textit{dynamically similar} if they are identical after normalization by certain parameters, inputs, and/or time \cite{ProfCartPole, Pratt_capture}. Dynamic similarity of two LIP models requires satisfaction of virtual kinematic constraints between the states of the two systems. One example is a single kinematic constraint that normalizes by scale,
\begin{equation}
\label{dynsimkinconst}
    \frac{\xi_{1}}{h_{1}} = \frac{\xi_{2}}{h_{2}}
\end{equation}
where $\xi:= x + \frac{\dot{x}}{\omega}$ is the divergent component of motion (DCM) of the LIPM \cite{DCM_intro}. Eq.\,(\ref{dynsimkinconst}) is satisfied by enforcing a corresponding dynamic constraint throughout the motion of both systems. The dynamic constraint is derived by differentiating both sides of Eq. (\ref{dynsimkinconst}) and normalizing by time,
\begin{equation}
\label{dynsimdynconst}
    \frac{\dot{\xi}_{1}}{h_{1}\omega_{1}} = \frac{\dot{\xi}_{2}}{h_{2}\omega_{2}}.
\end{equation}
\indent As shown in \cite{Prof_TRO}, satisfying Eq.\,(\ref{dynsimkinconst}) and Eq.\,(\ref{dynsimdynconst}) ensures normalized CoP equivalence between two non-passive LIP models. This is a highly desirable property for telelocomotion of bipedal robots as it enables synchronized stepping between human pilot and bipedal robot. 

\subsection{Whole-Body Human-Machine Interface}
\label{HIMsect}

The dynamic telelocomotion framework proposed in this work has unique HMI requirements (see Fig. \,ref{TelelocomotionConcept}). Mainly, it necessitates hardware capable of both motion capture and whole-body haptic feedback to a human pilot. The HMI must be able to measure human pilot CoM, feet position, net contact wrench, and apply large haptic forces to the pilot's torso. The design of such an interface has been studied in \cite{Prof_HMI, SunyuHMI, MartySATYRR}. 

\section{Human Whole-Body Dynamic Telelocomotion}
\label{CorePaper}

\begin{figure}[t]
\begin{center}
\includegraphics[width=1.0\linewidth]{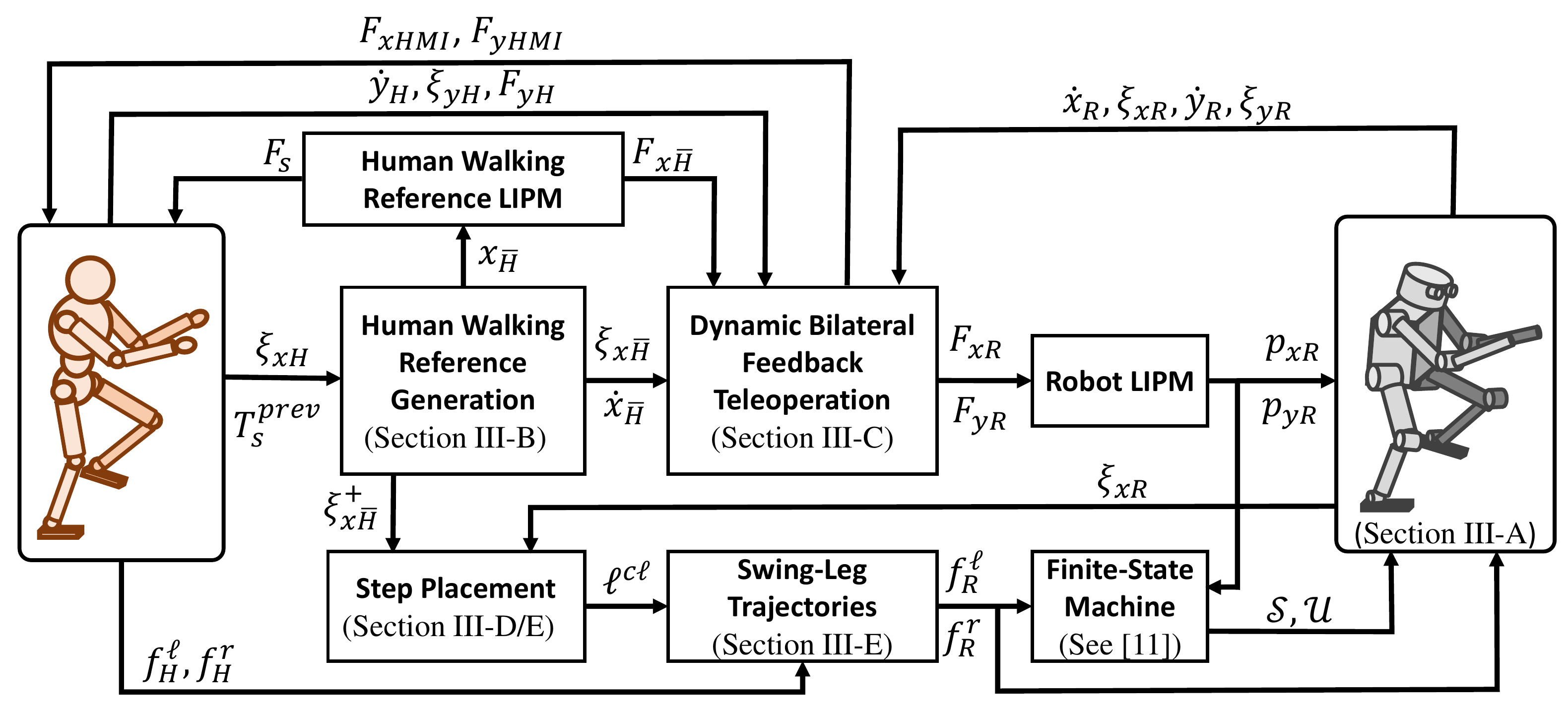}
\end{center}
\caption{Dynamic Telelocomotion Framework}
\label{block_diagram}
\end{figure}

The human pilot is responsible for producing a ``stable" reference gait to maintain balance of a teleoperated bipedal robot. To enforce the synchrony between the reference gait and the bipedal robot gait, the HMI continuously captures the movement and force generated by the pilot, and simultaneously applies large haptic forces on the pilot's torso such that the pilot perceives the misalignment in the gait synchrony.

There are two major challenges for sagittal plane walking; first, the human pilot is constrained to translate within a bounded HMI workspace, which inevitably restricts the gait of the pilot to mostly to stepping; second, the dynamics of human and robot evolve at different scales. To bridge the gap between two different walking behaviors, we introduce an interface; \textit{a virtual human walking reference model generated from human stepping}. This human proxy model serves as the reference for the bipedal robot to walk. Next, to dynamically synchronize the human walking reference and robot we enforce dynamic similarity by scaling the human walking reference to robot proportions. Finally, we couple three systems, robot, human walking reference, and human pilot via dynamic bilateral feedback teleoperation (DBFT). 

Fig. (\ref{block_diagram}) summarizes our dynamic telelocomotion framework with sub-components to be discussed in this section. For each system, bipedal locomotion is abstracted using the LIPM. Lastly, it should be noted that for the frontal plane dynamics, we adopt the methods from \cite{Prof_TRO}. 

\subsection{Reduced-Order Model Dynamics}
\label{robotmodel}

\begin{figure}[t!]
\begin{center}
\adjincludegraphics[trim={{0.0\width} {.61\height} {0.52\width} {0.0\height}},clip,width =.88\linewidth]{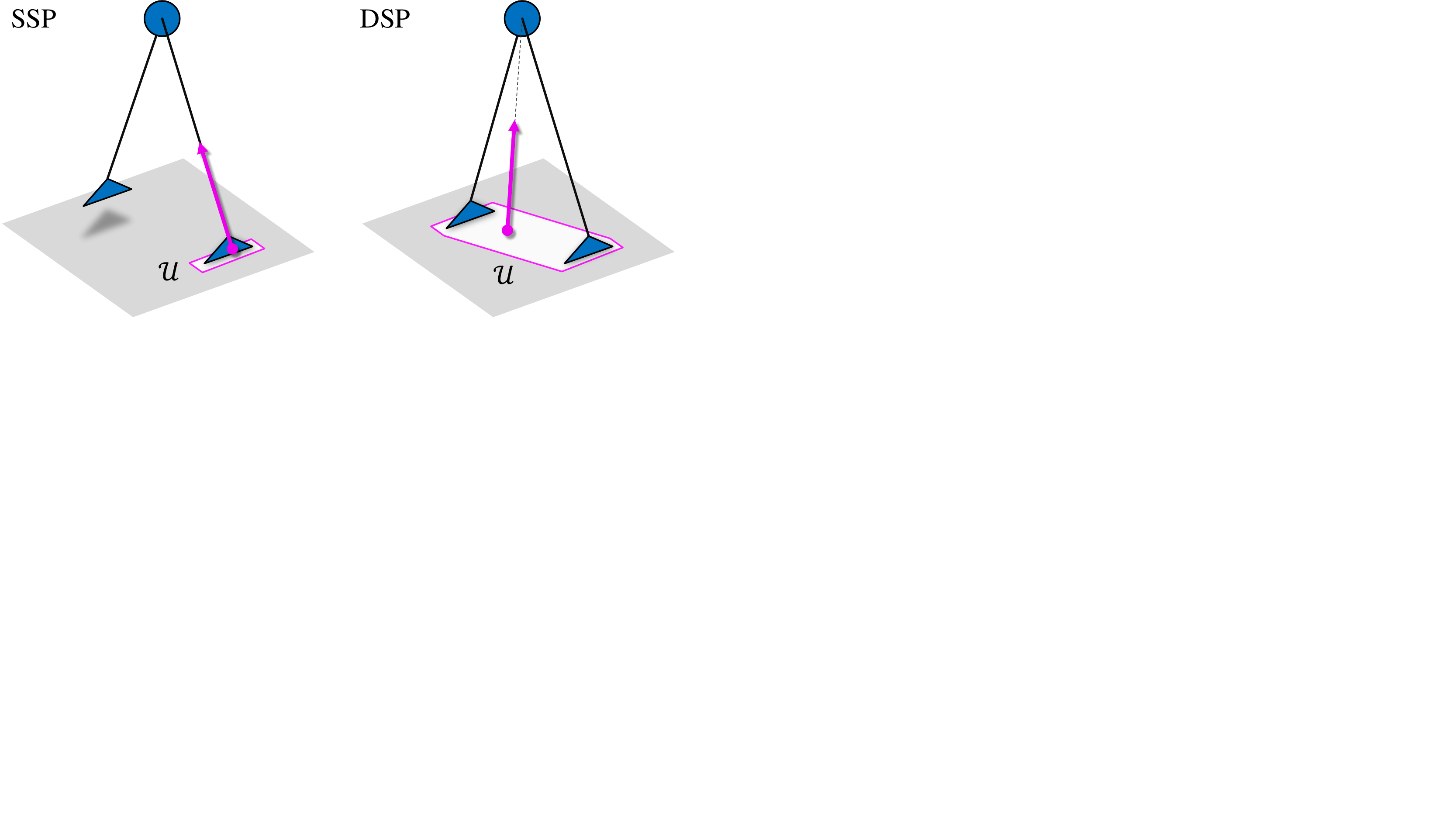}
\end{center}
\caption{Bipedal robot reduced-order model in the two domains of walking: SSP \& DSP.}
\label{2DLIPM_fig}    
\end{figure}

We consider a bipedal robot reduced-order model composed of two decoupled non-passive LIP models, massless support legs, and line feet (see Fig. \ref{2DLIPM_fig}). The hybrid model dynamics are single domain and defined as
\begin{equation}
\label{hybriddyn}
  \Sigma :
    \begin{cases}
      {\dot{\bm{x}} = f(\bm{x}, \bm{u})}, &t \not\in \mathcal{S} \\
        \bm{x}^{+} = \Delta ({\bm{x}^{-}}), &t \in \mathcal{S}
    \end{cases}  
    ,
\end{equation}
where $t$ is time, $\bm{x} = \left[x_{\sr}, \: \dot{x}_{\sr}, \:y_{\sr}, \:\dot{y}_{\sr} \right]^{\top} \in \mathbb{\R}^{4}$ is the vector of robot CoM states along the sagittal and frontal plane, and $\bm{u} = \left[p_{x\sr}, \:p_{y\sr} \right]^{\top} \in \mathcal{U}$ is the vector of robot CoP locations along the same planes. In order to simulate realistic walking dynamics we define the set of allowable control inputs, $\mathcal{U}$, as the area that spans the supporting feet, i.e., $
    \mathcal{U} = \{\bm{u} \in \mathbb{\R}^{2}: p_{x\sr}^{lb} \leq p_{x\sr} \leq p_{x\sr}^{ub}, p_{y\sr}^{lb} \leq p_{y\sr} \leq p_{y\sr}^{ub} \}$,
where the superscripts $lb$ and $ub$ denote lower and upper bounds, respectively. The guard $\mathcal{S}$ of the hybrid system is the transition $\Delta_{D\rightarrow S}$ from the double support phase (DSP) to the single support phase (SSP) of the robot. Such transitions are detected using the finite-state machine (FSM) from \cite{Prof_TRO}. The discrete transition maps the CoM x-position at the end of the current step to the beginning of the next step using the step length $\ell$. The reset map is given by
\begin{equation}
\label{discretemap}
  \Delta_{D\rightarrow S} :
    \begin{cases}
        x_{\sr}^{+} = x_{\sr}^{-} - \ell \\
        \dot{x}_{\sr}^{+} = \dot{x}_{\sr}^{-} \\
        y_{\sr}^{+} = y_{\sr}^{-} \\
        \dot{y}_{\sr}^{+} = \dot{y}_{\sr}^{-} 
    \end{cases}  
    .
\end{equation}
The continuous-time dynamics of the system are governed by Eq. (\ref{LIPeqmain}) along each plane. As seen in Eq.\,(\ref{discretemap}), walking is assumed to occur along the sagittal plane.

\subsection{Human Walking Reference LIPM Generation}
\label{humanwalkingrefLIPM}

\begin{figure}[t!]
\begin{center}
\adjincludegraphics[trim={{0.0\width} {.60\height} {0.45\width} {0.0\height}},clip,width =1\linewidth]{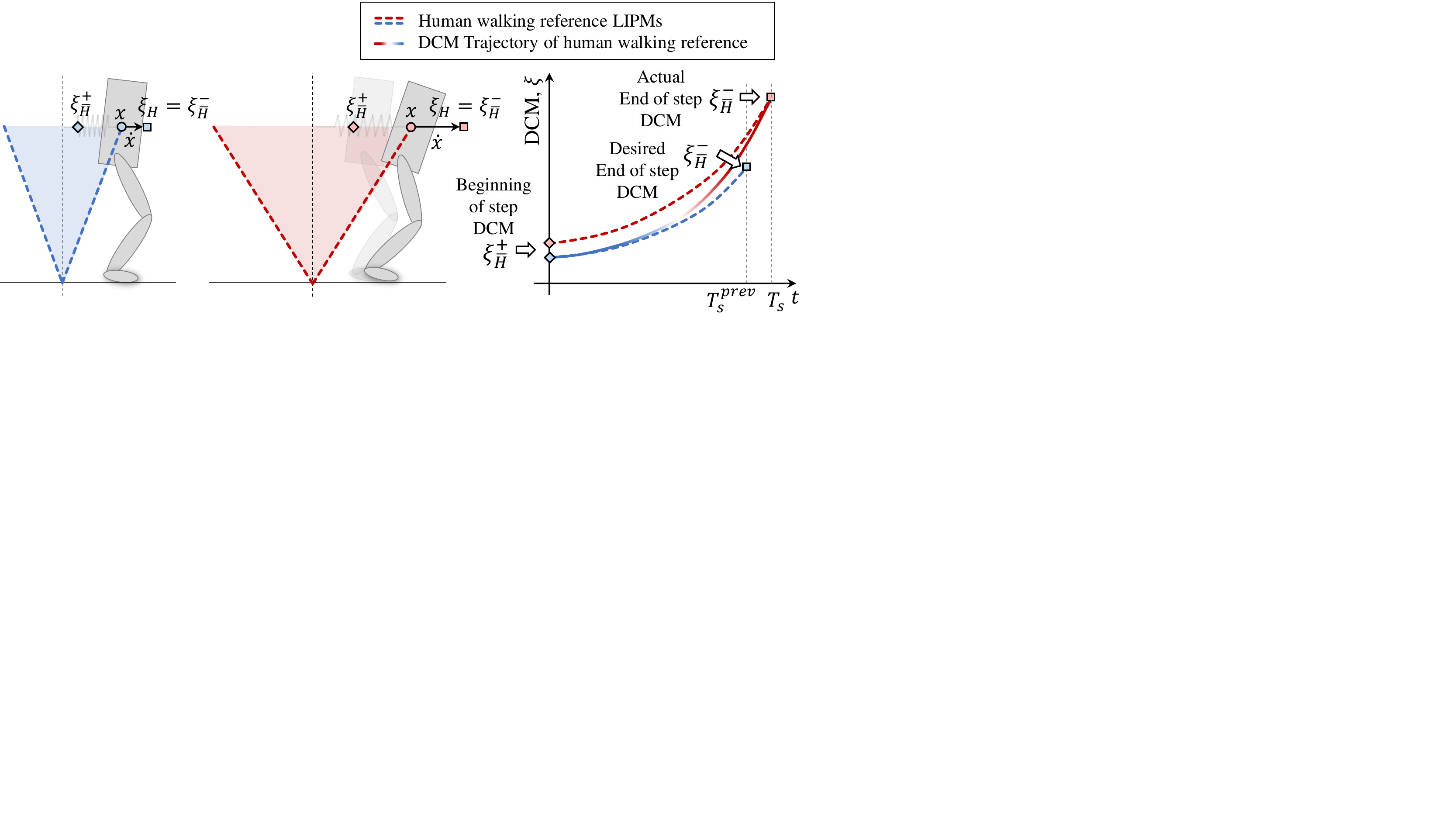}
\end{center}
\caption{A stable human walking reference trajectory is generated from human pilot stepping parameters, stepping frequency and DCM, at any given instant. As the pilot moves, the virtual human walking trajectory is continuously updated. The virtual spring modulates the contact force generated by the pilot.}
\label{endofstepDCMconcept} 
\end{figure}

In order to produce dynamic synchronization between human pilot stepping and bipedal robot walking, we introduce an interface, a human walking reference LIPM (HWR-LIPM) generated from the pilot’s stepping . One challenge in producing a gait (walking) from a dynamically distinct gait (stepping) is that we cannot predict human intent precisely, e.g., step placement or step timing. Hence, we infer from human motion a way to continuously adapt the HWR-LIPM state trajectory of the current step. Also, we hypothesize that two variables of gait, stepping frequency and DCM, can efficiently abstract walking and be utilized to produce a stable walking gait for the robot to track. Henceforth, the generation of the HWR-LIPM from human stepping is discussed (see Fig.\,\ref{endofstepDCMconcept}). 

The key idea to generate the HWR-LIPM is to assume, roughly speaking,
the human pilot’s \textit{intent of current desired walking behavior dictates the end-of-step stability of the HWR-LIPM}. Technically, we enforce that the measured human pilot DCM equals the end-of-step DCM of the HWR-LIPM. 

We assume that HWR-LIPM is passive and moves along a stable P1 orbit. As shown next, this means that the HWR-LIPM trajectory through a step can be back-calculated from assumed step time, $T_{s}$, and the desired end-of-step DCM, as in Fig.\,(\ref{endofstepDCMconcept}). Let us denote the superscripts $+$ and $-$ to describe LIPM states at the beginning-of-step and at the end-of-step, respectively. 
The subscripts $\sh$ and $\shbar$ correspond to the human LIPM (H-LIPM) and the HWR-LIPM, respectively. Parameters are assumed to be the same for both models. 

From hereon we derive the HWR-LIPM DCM trajectory. The end-of-step DCM of the walking reference is 
\begin{equation}
\label{endofstepdcm_walkingref}
    \xi_{x\shbar}^{-} = x_{\shbar}^{-} + \frac{\dot{x}_{\shbar}^{-}}{\omega_{\sh}}.
\end{equation}
To satisfy the necessary and sufficient conditions for stable P1 orbits the end-of-step CoM states are constrained to lie on the orbital line of characteristic with positive orbital slope, 
\begin{equation}
\label{orbitallinecondition}    
    \dot{x}_{\shbar}^{-} = \sigma_{1\sh}x_{\shbar}^{-},
\end{equation}
where $\sigma_{1\sh}$ is the orbital slope of the H-LIPM at the stepping frequency $\frac{1}{T_{s}}$ (ref. Eq.\,(\ref{orbitalslopeeq})). \textit{Now we apply the key idea, which is to substitute $\xi_{x\sh}$ for $\xi_{x\shbar}^{-}$}. This allows us to obtain end-of-step CoM states solely in terms of the human pilot stepping behavior by combining Eq. (\ref{endofstepdcm_walkingref}) and Eq. (\ref{orbitallinecondition}),
\begin{equation}
\label{endofstepCoM}    
    \begin{cases}
        x_{\shbar}^{-}(T_{s}, \xi_{x\sh}) = \xi_{x\sh} \slash 
        ( {1 + \frac{\sigma_{1\sh}}{\omega_{\sh}}}
        ), 
        \\
        \dot{x}_{\shbar}^{-}(T_{s}, \xi_{x\sh}) = \omega_{\sh}
        (\xi_{x\sh} - x_{\shbar}^{-} ),
    \end{cases}
\end{equation}
Now, since the CoM states evolve along a stable P1 orbit, their beginning-of-step values must be
\begin{equation}
\label{begofstepCoM}    
    \begin{cases}
        x_{\shbar}^{+} = -x_{\shbar}^{-}, \\ 
        \dot{x}_{\shbar}^{+} = \dot{x}_{\shbar}^{-},   
    \end{cases}
\end{equation}
such that they lie on the orbital line of characteristic with negative orbital slope. The result from Eq.\,(\ref{endofstepCoM}) and Eq.\,(\ref{begofstepCoM}) is that in this work $\xi_{x\shbar}^{+}:= f(T_{s}, \xi_{x\sh})$. Re-writing Eq.\,(\ref{passiveLIPsol}) in terms of a DCM change of coordinates and initializing the HWR-LIPM with $\xi_{x\shbar}^{+}$ yields a closed-form expression for the DCM trajectory of the HWR-LIPM,
\begin{equation}
\label{DCMtrajclosedform}    
    \xi_{x\shbar}(t) = \xi_{x\shbar}^{+}e^{\omega t},
\end{equation}
where $t \in [0, T_{s}]$. Eq.\,(\ref{DCMtrajclosedform}) implies that the human pilot can update the HWR-LIPM trajectory by modulating $\xi_{x\shbar}^{+}$ via their stepping frequency and instantaneous DCM as illustrated in Fig. (\ref{endofstepDCMconcept}).  

\subsection{Dynamic Bilateral Feedback Teleoperation}
\label{telelocomotionlaw}

Let us consider the sagittal plane CoM dynamics of these three coupled systems: robot (subscript $\sr$), human walking reference, and human pilot. The equations of motion for each system can be written as
%
\begin{align}
\label{robotCoMdyn}
    &\textrm{R-LIPM: }&&m_{\textsc{r}}\ddot{x}_{\sr} = F_{x\sr}^{\sff} + F_{x\sr}^{\sfb} + F_{ext},
\\
\label{humanrefCoMdyn}
    &\textrm{HWR-LIPM: }&&m_{\sh}\ddot{x}_{\shbar} = F_{x\shbar} + F_{x\shmi},
\\
\label{humanCoMdyn}
    &\textrm{H-LIPM: }&&m_{\sh}\ddot{x}_{\sh} = F_{x\sh} - F_{s} + F_{x\shmi},
\end{align}
where $F_{x\sr}^{\sff}$ and $F_{x\sr}^{\sfb}$ are feedforward and feedback components of the robot's controller, respectively, $F_{ext}$ is an external disturbance applied to the robot, $F_{x\shbar}$ is the human walking reference contact force, $F_{x\shmi}$ is the haptic feedback force applied to the human pilot, $F_{x\sh}$ is the contact force generated by the human pilot on the force plate, and $F_{s}$ is the virtual spring force generated by the HMI to regulate human pilot movement. Consequently, the interplay between these dynamics is carried out by several forces.

First, the robot control forces, $F_{x\sr}^{\sff}$ and $F_{x\sr}^{\sfb}$, and human feedback force, $F_{x\shmi}$, couple the human walking reference and robot, by enforcing dynamic similarity\,\cite{Prof_TRO} between the HWR-LIPM and R-LIPM;
\begin{gather}
    F_{x\sr}^{\sff} = \frac{m_{\sr}h_{\sr}\omega_{\sr}^{2}}{m_{\sh}h_{\sh}\omega_{\sh}^{2}}F_{x\shbar}, \label{robotfeedforward}
    \\
    F_{x\sr}^{\sfb} = K_{x}
    \Big(
    \frac{\xi_{x\shbar}}{h_{\sh}} - \frac{\xi_{x\sr}}{h_{\sr}}
    \Big)
    ,\label{robotfeedback}
    \\
    \begin{aligned}
    F_{x\shmi} = m_{\sh}h_{\sh}\omega_{\sh}^2
    \Big(
    \frac{\dot{x}_{\sr}}{h_{\sr}\omega_{\sr}} \!-\! \frac{\dot{x}_{\shbar}}{h_{\sh}\omega_{\sh}}
    \Big) \!+\! 
    \frac{m_{\sh}h_{\sh}\omega_{\sh}^{2}}{m_{\sr}h_{\sr}\omega_{\sr}^2}F_{ext},
    \end{aligned}
    \label{HMIforce}
\end{gather}
where $F_{x\shbar}$ is given by Eq.\,(\ref{nonpassiveLIPforce}) with $p_{x\shbar} = 0$, $K_{x}$ is a feedback gain similar to in \cite{Prof_TRO}, and $x_{\shbar}$ and $\dot{x}_{\shbar}$ are given by Eq.\,(\ref{passiveLIPsol}) with initial conditions from Eq.\,(\ref{begofstepCoM}).


Next, the walking reference and the pilot are coupled such that the human is forced to apply a ground contact force $F_{x\sh}$ consistent with the walking reference. To achieve this, the virtual spring force $F_s$ serves two purposes. First, it works against $F_{x\sh}$ so that the pilot has to apply effort to achieve forward robot walking. This makes generating a given walking reference more intuitive. We set $F_{s} = F_{x\shbar}^{+}$, which yields   
\begin{equation}
\label{virtualspringeq}
    F_{s} = m_{\sh}\omega_{\sh}^2 x_{\shbar}^{+}.
\end{equation}
Second, it enables, under steady-state teleoperation conditions ($\dot{x}_{\sh} \approx 0$), any perceived change in the human pilot CoM dynamics at touch-down to be attributed to the misalignment between R-LIPM and HWR-LIPM. This is because the force that enforces dynamic similarity for the walking reference is also applied to the pilot. Thus, indirectly coupling the pilot and robot.

\subsection{Step Placement Strategy}
\label{stepplacement}

The robot's swing-leg touch-down is dictated by human stepping behavior. More specifically, the transition $\Delta_{D\rightarrow S}$ is governed by the frontal plane DBFT dynamics, i.e., the normalized CoP synchronization of human pilot and robot in the frontal plane. Thus, for the step placement to be consistent with the coupled dynamics, we enforce dynamic similarity between the HWR-LIPM and R-LIPM at the beginning of the next step. In other words, normalized DCM tracking should be invariant to the reset map from Eq. (\ref{discretemap}), which results in the following kinematic constraint:
\begin{equation}
\label{dynsimkinconstbegofstep}
    \frac{\xi_{x\shbar}^{+}}{h_{\sh}} = \frac{\xi_{x\sr}^{+}}{h_{\sr}}.
\end{equation}
Now we can derive an expression for step length, $\ell$, that enforces Eq.\,(\ref{dynsimkinconstbegofstep}) by using the reset map and the definition of the beginning-of-step DCM. After some simplification the resulting expression is given by
\begin{equation}
\label{steplengtheq}
    \ell = \xi_{x\sr}^{-} - \xi_{x\shbar}^{+}\left (\frac{h_{\sr}}{h_{\sh}} \right ).
\end{equation}
Due to robot swing-leg touch-down behavior, at any time, $\xi_{x\sr} \rightarrow \xi_{x\sr}^{-}$, could occur. Thus, Eq. (\ref{steplengtheq}) becomes a control feedback law that minimizes the error in dynamic similarity between the HWR-LIPM and R-LIPM at a step transition.

\subsection{Implementation Details}
\label{implementation}

To allow the human pilot to vary step frequency, we assume the current step will take the same time as the previous step ($T_{s} = T_{s}^{prev}$). This is a heuristic used because we cannot know when the current step comes to an end. If $T_{s}^{curr} > T_{s}^{prev}$, we track $\xi_{\sh}$ and elongate the human walking reference trajectory. This is shown in Fig.\,(\ref{endofstepDCMconcept}), where the red dotted line is elongated further than blue dotted line. Otherwise, the desired step may not be taken and this acts as a disturbance to the walking controller. We assume since humans step mostly periodically that these disturbances are small and the human pilot can stabilize the robot walking dynamics smoothly via DBFT. 

Rapidly estimating the human DCM could be difficult because the DCM includes a CoM velocity term which is usually noisy. We use the human CoM, $x_{\sh}$, as a surrogate for the DCM of the human, $\xi_{x\sh}$, past a certain dead-band value, $x_{\sh}^{db}$. This heuristic is a reasonable choice because human pilot mostly steps-in-place during teleoperation. We tune $x_{\sh}^{db}$ to improve sagittal plane balancing of the robot, especially when $\dot{x}_{\sr} \approx 0$ is desired.

Although the single domain hybrid system assumption is generally valid, the effectiveness of step placement strategy (Eq.\,(\ref{dynsimkinconstbegofstep})) degrades due to unaccounted $x_{\sr}^{+}$ displacement during DSP. As in \cite{HLIP_AMBER}, we add an approximate feedforward step-length 
to obtain a final closed-loop step placement law,
\begin{equation}
\label{steplengtheqfinal}
    \ell^{c\ell} = \xi_{x\sr} - \xi_{x\shbar}^{+}\left (\frac{h_{\sr}}{h_{\sh}} \right ) + \dot{x}_{\sr}T_{DSP}.
\end{equation}
where $T_{DSP}$ is the assumed time duration of the DSP. 

Finally, robot swing-leg trajectories $f_{\sr}^{r}$ and $f_{\sr}^{\ell}$ are composed of normalized human frontal plane swing-leg position, $f_{\sh}^{r}$ and $f_{\sh}^{\ell}$, and a smooth mapping of swing-to-stance foot as in \cite{gong2021angular} implementing the step length from Eq.\,(\ref{steplengtheqfinal}). Swing-leg dynamics are ignored in this work.

\section{Results}
\label{resultsexp}

\begin{figure}[t]
\begin{center}
\includegraphics[width=1.0\linewidth]{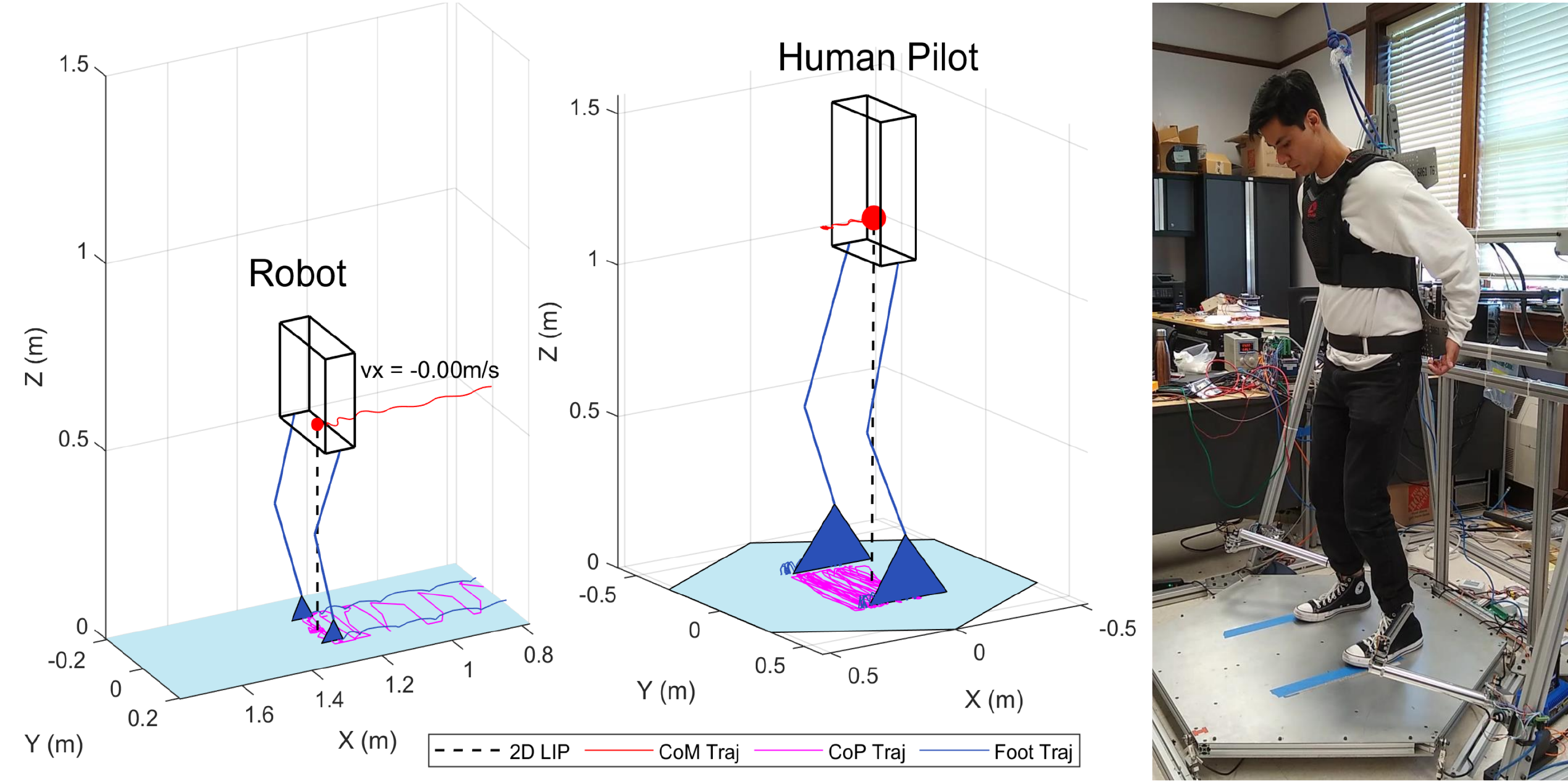}
\end{center}
\caption{The simulated robot (left), human motion capture (middle), and experimental set-up (right).}
\label{sim_exp}
\end{figure}

\begin{table}[]
\caption{Reduced-Order Model Parameters}
\label{paramtable}
\begin{tabular}{llllll}
\hline
Parameter         & Symbol & Human & Tello & Units & Ratio \\ 
\hline
Mass              & $m$       &  75.0 &     20.2    & kg         & 3.71      \\
CoM Height        & $h$       &  1.20 &     0.55    & kg         & 2.18      \\
Natural Frequency & $\omega$  &  2.85 &     4.22    & $s^{-1}$   & 0.68      \\
Foot Length       & $l_{f}$   &  0.31 &     0.06    & m          & 5.17      \\
Foot Width        & $w_{f}$   &  0.11 &     0.03    & m          & 3.67     
\end{tabular}
\end{table}

In this work, using the HMI, a trained human pilot leverages their intelligence and sensorimotor skills to control the reduced-order dynamics of a bipedal robot using our whole-body dynamic telelocomotion framework (see Fig. \ref{block_diagram}). The pilot completes three different bipedal locomotion tasks without any visual feedback of the simulated robot. 

The first task is to achieve synchronized stepping-in-place while maintaining robot balance in the frontal and sagittal plane. This is meant to validate if the proposed framework can reproduce the results from \cite{Prof_TRO} and regulate sagittal plane robot CoM velocity near zero ($\dot{x}_{\sr} \approx 0$). The second task is to showcase the major contribution of our work, which is to achieve, to the best of our knowledge, the \textit{first synchronized bipedal walking between human and humanoid robot}. More specifically, we want to observe whether the human pilot can  control robot forward walking from a standstill and bring the robot to a complete stop after walking for at least $1$\,m and $10$\,steps. Lastly, the final task is the classic ``kick" to the robot CoM. We apply a known horizontal $30$\,N force, for approximately $0.3$\,s, to the robot CoM which scales by weight to $111$\,N for the human and is channeled to the pilot via DBFT. The goal is to see if the human pilot can use their natural sensorimotor  skills (e.g. reflexes) to stabilize the robot and prevent it from falling. 

\subsection{Experimental Setup}

For each task discussed above, the human pilot is only instructed when to commence teleoperation. As mentioned, the pilot has no visual feedback of the simulated robot.
The simulation of reduced-order robot dynamics, human whole-body dynamic telelocomotion framework, and HMI interface were computed at $1$\,kHz in a real-time computer (cRIO-9082, National Instruments). 
The hardware used in these experiments is based on the HMI in \cite{MartySATYRR} with an addition of a lower-body motion capture device that measures feet position, similar to in \cite{Prof_HMI}. The parameters of the reduced-order robot model correspond to those of the humanoid Tello \cite{TELLOpaper}, which are compared to the parameters of the human pilot in Table\,\ref{paramtable}. Fig.\,(\ref{sim_exp}) shows the experimental set-up as well as a visualization of the the human pilot motion capture data and the robot reduced-order model simulation results.


\subsection{Results \& Discussion}

\begin{figure}[t]
\begin{center}
\includegraphics[width=1.0\linewidth]{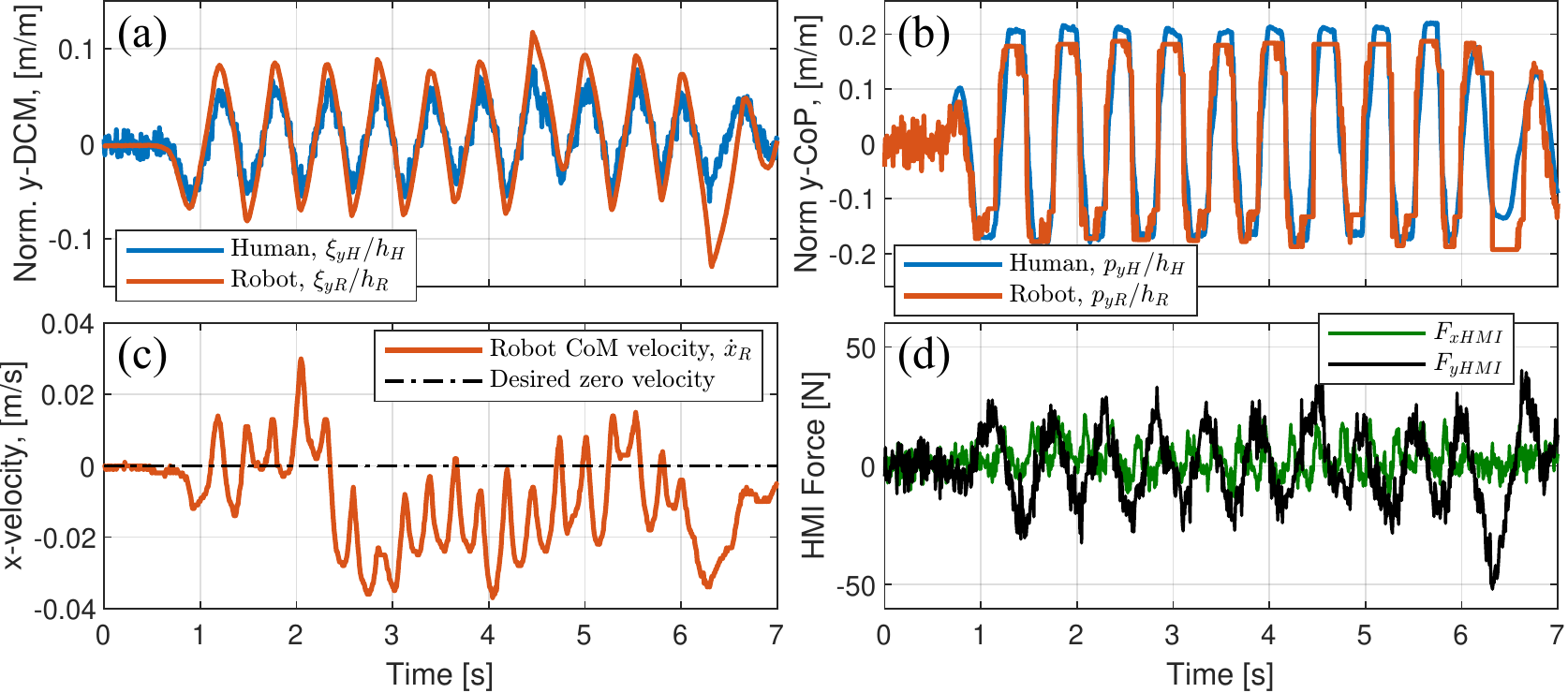}
\end{center}
\caption{Stepping-in-place experiment results. (a) normalized DCM, (b) normalized CoP, (c) robot velocity regulation, (d) feedback force on the pilot. }
\label{stepping-in-place}
\end{figure}

In these experiments, a human pilot successfully leveraged their intelligence to plan desired locomotion behaviors (e.g. step-in-place, walk) and/or their sensorimotor skills to maintain balance (e.g. reject a disturbance) for the robot. 

First, the robot stepped-in-place for approximately $6$\,s ($20$ steps), synchronized with the human. Our framework was able to reproduce the results from \cite{Prof_TRO, ProfScienceRob}, and show that the dependency between normalized CoP synchronization (Fig.\,\ref{stepping-in-place}b) and normalized DCM tracking (Fig.\,\ref{stepping-in-place}a) is the result of dynamic similarity. Moreover, the proposed method offers an improvement to \cite{Prof_TRO, ProfScienceRob}, as the human pilot now also balances the robot in the sagittal plane, regulating CoM velocity near zero (Fig.\,\ref{stepping-in-place}c). Thanks to good normalized DCM tracking, the haptic feedback forces to the human pilot remain stable and consistent (Fig.\,\ref{stepping-in-place}d). 
 
\begin{figure}[t]
\begin{center}
\includegraphics[width=1.0\linewidth]{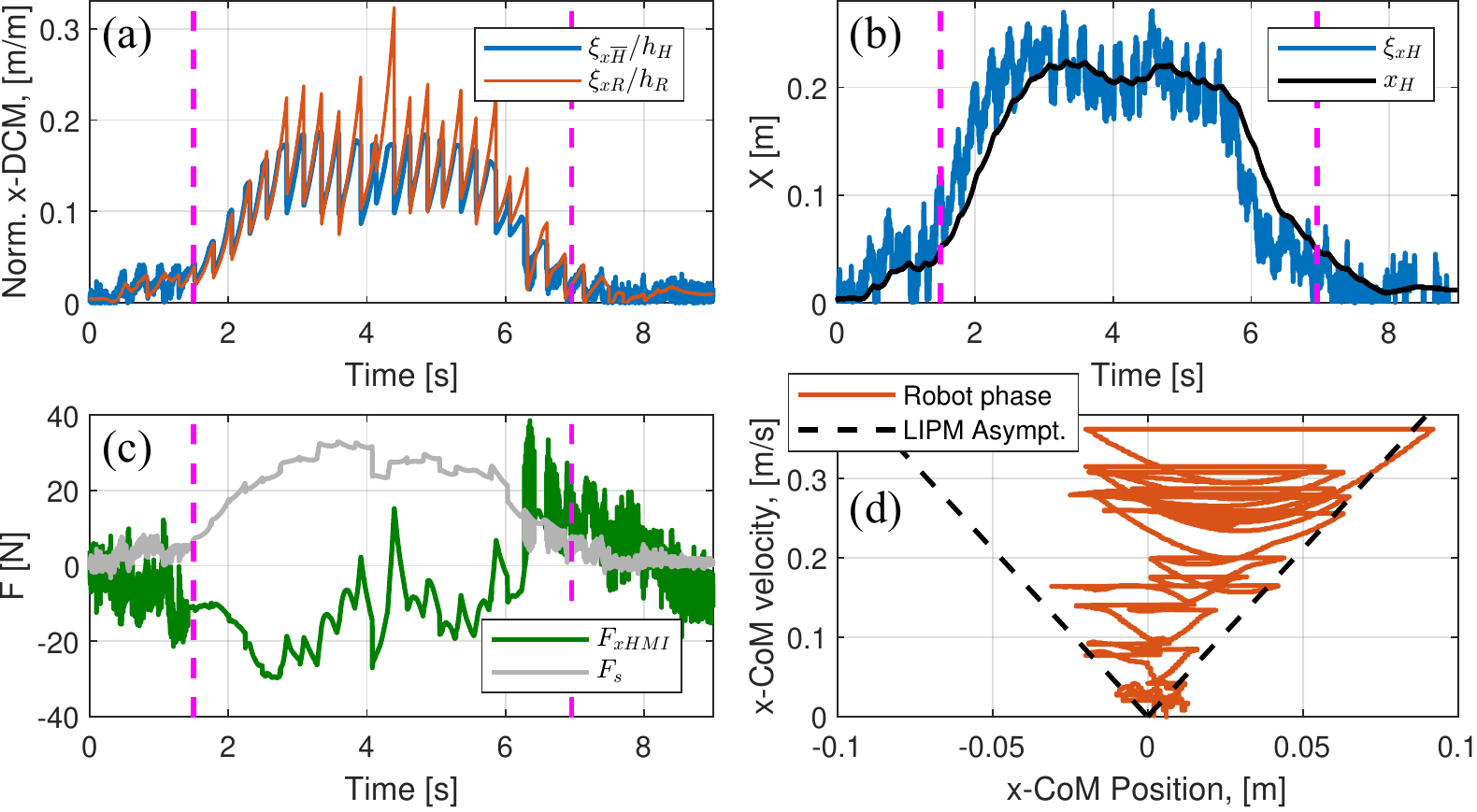}
\end{center}
\caption{Forward walking experiment results. (a) normalized DCM, (b) trajectory of the pilot's states, (c) forces on the pilot, (d) phase diagram of robot locomotion.}
\label{walking-forward-experiment}
\end{figure}

Second, \textit{The human pilot, using our dynamic telelocomotion framework, was able to initiate, control, and terminate bipedal robot walking without any visual feedback.}. To the best of our knowledge, this has not been shown in any previous work. The robot walked $1.28$\,m (over $10$\,steps) and reached a top speed of $0.36$\,m/s, which equates to 40\% of average human walking speed. The human walking reference generated by the pilot (Fig.\,\ref{walking-forward-experiment}b) was tracked by the robot controller aided by the step placement strategy that maintains dynamic similarity at step transitions (Fig.\,\ref{walking-forward-experiment}a). The virtual spring force and haptic feedback force (Fig.\,\ref{walking-forward-experiment}c) help the pilot generate and update the walking reference, respectively. For example, around the $4$\,s mark, although the normalized DCM of the robot spikes, the pilot was able to bring the system back to balance. The phase portrait of the robot CoM dynamics (Fig.\,\ref{walking-forward-experiment}d) highlights the prowess of human capability. Despite the apparent non-periodicity of the robot gait due to assumed step timing, the human pilot adapts and regulates the telelocomotion and keeps the robot walking. In this experiment, the average error in assumed step timing was $34$\,ms with a standard deviation of $28$\,ms. Based on this result, we claim the efficacy of the proposed framework. \\
\begin{figure}[t]
\begin{center}
\includegraphics[width=1.0\linewidth]{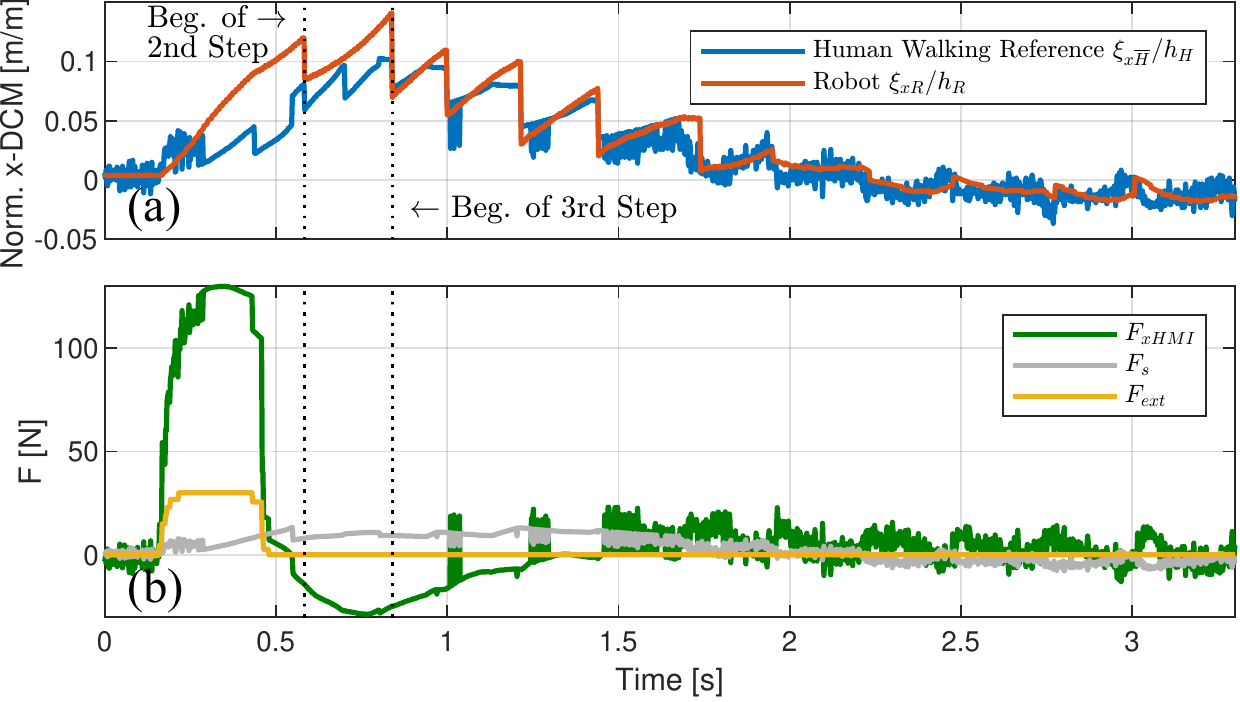}
\end{center}
\caption{Disturbance ($.2$\,s $\sim$ $.45$\,s) rejection experiment results. (a) normalized x-DCM, (b) forces on the pilot ($F_{x\shmi}$, $F_s$) and disturbance on the robot ($F_{ext}$).} 
\label{disturbance-rejection}
\end{figure}
$\,\,$ As to the disturbance rejection experiment, the human pilot, feeling a large push, instinctively takes a step forward before stabilizing their gait. The robot and human pilot resume synchrony at the 3rd step after the disturbance (Fig.\,\ref{disturbance-rejection}a). This means that it took only two steps for the human pilot to re-stabilize the robot dynamics. The result implies that the natural balancing strategy of a human, which can reject a sizeable $130$\,N peak disturbance (Fig.\,\ref{disturbance-rejection}b), can be successfully transferred to a robot.

Although this work is a promising step towards harnessing human prowess to realize humanoid robot performance capable of meeting today's demand for these robots, there are some limitations. First, the simulated robot uses ideal reduced-order model dynamics which assume constant height walking and no moment around the CoM. Bipedal robots often regulate angular dynamics to extend their CoP \cite{ProfScienceRob}. Another limitation is that the teleoperation requires extensive training to achieve the demonstrated performance. Other limitations include robot walking speed not scaling exactly to the walking speed of the average human and incorrect assumption on step timing. 

\section{CONCLUSION}

To conclude, we present a human whole-body dynamic telelocomotion framework that achieves the first synchronization of human stepping with bipedal robot walking. To accomplish this, we introduce a method to construct a human walking reference LIPM via human stepping frequency and DCM. This walking reference is then synchronized with the robot using dynamic bilateral feedback teleoperation as well as a consistent robot step placement strategy. Experiments demonstrated that the proposed framework enables human pilot to control stepping and walking of a simulated bipedal robot and transfer human balancing strategy to a robot. 

Future work will test the proposed framework on the humanoid Tello. We will adopt state-of-the-art legged locomotion controller architectures \cite{MITHUMANOID} \cite{donghyun_wbc} consisting of a planner and a tracking controller by using human intelligence as the planner and using a richer reduced-order model and whole-body controller.

\bibliographystyle{unsrt}
\bibliography{main.bib}

\begin{thebibliography}{10}

\bibitem{scott_atlas}
Scott Kuindersma, Robin Deits, Maurice Fallon, Andr{\'e}s Valenzuela, Hongkai
  Dai, Frank Permenter, Twan Koolen, Pat Marion, and Russ Tedrake.
\newblock Optimization-based locomotion planning, estimation, and control
  design for the atlas humanoid robot.
\newblock {\em Autonomous Robots}, 40(3):429--455, March 2016.

\bibitem{grizzle_ames}
Jessy~W. Grizzle, Christine Chevallereau, Ryan~W. Sinnet, and Aaron~D. Ames.
\newblock Models, feedback control, and open problems of 3d bipedal robotic
  walking.
\newblock {\em Automatica}, 50(8):1955--1988, 2014.

\bibitem{MITHUMANOID}
Matthew Chignoli, Donghyun Kim, Elijah Stanger-Jones, and Sangbae Kim.
\newblock The mit humanoid robot: Design, motion planning, and control for
  acrobatic behaviors.
\newblock In {\em 2020 IEEE-RAS 20th International Conference on Humanoid
  Robots (Humanoids)}, pages 1--8, 2021.

\bibitem{donghyun_wbc}
Donghyun Kim, Steven~Jens Jorgensen, Jaemin Lee, Junhyeok Ahn, Jianwen Luo, and
  Luis Sentis.
\newblock Dynamic locomotion for passive-ankle biped robots and humanoids using
  whole-body locomotion control.
\newblock {\em The International Journal of Robotics Research}, 39(8):936--956,
  2020.

\bibitem{wholebody_retarget_Darvish}
Kourosh Darvish, Yeshasvi Tirupachuri, Giulio Romualdi, Lorenzo Rapetti, Diego
  Ferigo, Francisco Javier~Andrade Chavez, and Daniele Pucci.
\newblock Whole-body geometric retargeting for humanoid robots.
\newblock In {\em 2019 IEEE-RAS 19th International Conference on Humanoid
  Robots (Humanoids)}, pages 679--686, 2019.

\bibitem{whole_body_movement_opt_retarget}
Waldez Gomes, Vishnu Radhakrishnan, Luigi Penco, Valerio Modugno, Jean-Baptiste
  Mouret, and Serena Ivaldi.
\newblock Humanoid whole-body movement optimization from retargeted human
  motions.
\newblock In {\em 2019 IEEE-RAS 19th International Conference on Humanoid
  Robots (Humanoids)}, pages 178--185, 2019.

\bibitem{whole_body_master_slave_locomotion}
Yasuhiro Ishiguro, Kunio Kojima, Fumihito Sugai, Shunichi Nozawa, Yohei
  Kakiuchi, Kei Okada, and Masayuki Inaba.
\newblock Bipedal oriented whole body master-slave system for dynamic secured
  locomotion with lip safety constraints.
\newblock In {\em 2017 IEEE/RSJ International Conference on Intelligent Robots
  and Systems (IROS)}, pages 376--382, 2017.

\bibitem{whole_body_master_slave_locomotion2}
Yasuhiro Ishiguro, Kunio Kojima, Fumihito Sugai, Shunichi Nozawa, Yohei
  Kakiuchi, Kei Okada, and Masayuki Inaba.
\newblock High speed whole body dynamic motion experiment with real time
  master-slave humanoid robot system.
\newblock In {\em 2018 IEEE International Conference on Robotics and Automation
  (ICRA)}, pages 5835--5841, 2018.

\bibitem{balancing_feedback_vibrotactile}
Anais Brygo, Ioannis Sarakoglou, Nadia Garcia-Hernandez, and Nikolaos
  Tsagarakis.
\newblock Humanoid robot teleoperation with vibrotactile based balancing
  feedback.
\newblock In Malika Auvray and Christian Duriez, editors, {\em Haptics:
  Neuroscience, Devices, Modeling, and Applications}, pages 266--275, Berlin,
  Heidelberg, 2014. Springer Berlin Heidelberg.

\bibitem{full_body_interface_compliant_robot_interaction}
Luka Peternel and Jan Babič.
\newblock Learning of compliant human–robot interaction using full-body
  haptic interface.
\newblock {\em Advanced Robotics}, 27(13):1003--1012, 2013.

\bibitem{Prof_TRO}
Joao Ramos and Sangbae Kim.
\newblock Humanoid dynamic synchronization through whole-body bilateral
  feedback teleoperation.
\newblock {\em IEEE Transactions on Robotics}, 34(4):953--965, 2018.

\bibitem{ProfScienceRob}
Joao Ramos and Sangbae Kim.
\newblock Dynamic locomotion synchronization of bipedal robot and human
  operator via bilateral feedback teleoperation.
\newblock {\em Science Robotics}, 4(35):eaav4282, 2019.

\bibitem{LittleHermes_CoDesign}
Joao Ramos, Benjamin Katz, Meng Yee~Michael Chuah, and Sangbae Kim.
\newblock Facilitating model-based control through software-hardware co-design.
\newblock In {\em 2018 IEEE International Conference on Robotics and Automation
  (ICRA)}, pages 566--572, 2018.

\bibitem{ProfCartPole}
Joao Ramos and Sangbae Kim.
\newblock Dynamic bilateral teleoperation of the cart-pole: A study toward the
  synchronization of human operator and legged robot.
\newblock {\em IEEE Robotics and Automation Letters}, 3(4):3293--3299, 2018.

\bibitem{Pratt_capture}
Jerry Pratt, John Carff, Sergey Drakunov, and Ambarish Goswami.
\newblock Capture point: A step toward humanoid push recovery.
\newblock pages 200 -- 207, 01 2007.

\bibitem{SunyuHMI}
Sunyu Wang and Joao Ramos.
\newblock Dynamic locomotion teleoperation of a reduced model of a wheeled
  humanoid robot using a whole-body human-machine interface.
\newblock {\em IEEE Robotics and Automation Letters}, 7(2):1872--1879, 2022.

\bibitem{HLIP_AMBER}
Xiaobin Xiong and Aaron~D. Ames.
\newblock Orbit characterization, stabilization and composition on 3d
  underactuated bipedal walking via hybrid passive linear inverted pendulum
  model.
\newblock In {\em 2019 IEEE/RSJ International Conference on Intelligent Robots
  and Systems (IROS)}, pages 4644--4651, 2019.

\bibitem{gong2021angular}
Yukai Gong and Jessy Grizzle.
\newblock Angular momentum about the contact point for control of bipedal
  locomotion: Validation in a lip-based controller, 2021.

\bibitem{DCM_intro}
Johannes Englsberger, Christian Ott, and Alin Albu-Schäffer.
\newblock Three-dimensional bipedal walking control based on divergent
  component of motion.
\newblock {\em IEEE Transactions on Robotics}, 31(2):355--368, 2015.

\bibitem{Prof_HMI}
Joao Ramos, Albert Wang, Wyatt Ubellacker, John Mayo, and Sangbae Kim.
\newblock A balance feedback interface for whole-body teleoperation of a
  humanoid robot and implementation in the hermes system.
\newblock In {\em 2015 IEEE-RAS 15th International Conference on Humanoid
  Robots (Humanoids)}, pages 844--850, 2015.

\bibitem{MartySATYRR}
Amartya Purushottam, Yeongtae Jung, Kevin Murphy, Donghoon Baek, and Jo{\~a}o
  Ramos.
\newblock Hands-free telelocomotion of a wheeled humanoid toward dynamic mobile
  manipulation via teleoperation.
\newblock {\em ArXiv}, abs/2203.03558, 2022.

\bibitem{TELLOpaper}
Youngwoo Sim and Joao Ramos.
\newblock Tello leg: The study of design principles and metrics for dynamic
  humanoid robots.
\newblock {\em IEEE Robotics and Automation Letters}, 7(4):9318--9325, 2022.

\end{thebibliography}

\end{document}